\documentclass[preprint,12pt,authoryear]{elsarticle}
\usepackage{amssymb}
\usepackage{amsmath}
\usepackage{float}
\restylefloat{table}
\usepackage{adjustbox}
\usepackage{wrapfig}
\usepackage{multirow}
\usepackage{xcolor}

\journal{IET Computer Vision}
\begin{document}
\begin{frontmatter}

\title{OmDet: Large-scale Vision-Language Multi-dataset Pre-training with Multimodal Detection Network}

\author[1] {Tiancheng Zhao}

\author[2] {Peng Liu}
\affiliation[2]{organization={Linker Technology Research},
            city={Hangzhou},
            postcode={310005}, 
            state={Zhejiang},
            country={China}}
\author[1] {Kyusong Lee}

\affiliation[1]{organization={Binjiang Institute of Zhejiang University},
            city={Hangzhou},
            postcode={310053}, 
            state={Zhejiang},
            country={China}}

\begin{abstract}
The advancement of object detection (OD) in open-vocabulary and open-world scenarios is a critical challenge in computer vision. This work introduces OmDet, a novel language-aware object detection architecture, and an innovative training mechanism that harnesses continual learning and multi-dataset vision-language pre-training. Leveraging natural language as a universal knowledge representation, OmDet accumulates a "visual vocabulary" from diverse datasets, unifying the task as a language-conditioned detection framework. Our multimodal detection network (MDN) overcomes the challenges of multi-dataset joint training and generalizes to numerous training datasets without manual label taxonomy merging. We demonstrate superior performance of OmDet over strong baselines in object detection in the wild, open-vocabulary detection, and phrase grounding, achieving state-of-the-art results. Ablation studies reveal the impact of scaling the pre-training visual vocabulary, indicating a promising direction for further expansion to larger datasets. The effectiveness of our deep fusion approach is underscored by its ability to learn jointly from multiple datasets, enhancing performance through knowledge sharing.

\end{abstract}


\end{frontmatter}

\section{Introduction}
Object detection (OD) is one of the monumental tasks in computer vision (CV). Classical OD research has been focusing on improving the detector network to achieve higher accuracy with lower latency~\citep{ren2015faster,redmon2016you,liu2016ssd,zhou2019objects}. In the last decade, numerous novel OD architectures have been developed, including well-known two-stage methods, e.g., Faster-RCNN~\citep{ren2015faster}, one-stage methods, e.g., Yolo, SSD, and CenterNet~\citep{redmon2016you,liu2016ssd,zhou2019objects}, and recent end-to-end methods e.g., DETR and Sparse-RCNN~\citep{carion2020end,sun2021sparse}. However, although the state-of-the-art OD model can achieve over 60 AP on COCO~\citep{lin2014microsoft}, classical OD systems all share one major limitation, i.e., they cannot generalize to object types that are not included in the pre-defined label set, such as 80 classes from COCO. This weakness limits OD such that it can only be applied to domains with known targets, and cannot be used in more challenging open-world domains, such as robotics, augmented reality, and embodied agents, that demand the system to detect any type of objects on the fly as the human users request.

On the other hand, Vision-Language Pre-training (VLP) has rapidly progressed~\citep{li2020oscar,radford2021clip,li2021align,kim2021vilt}, thanks to the emergence of multimodal transformers~\citep{vaswani2017attention} and the availability of large paired image-text corpora~\citep{sharma2018conceptual,changpinyo2021conceptual}. By learning image-to-text matching from massive multimodal datasets, many proposed VLP models have helped to achieve the state-of-the-art performance of a variety of downstream multimodal tasks, ranging from visual QA~\citep{lu2019vilbert}, cross-modal retrieval~\citep{lu2021visualsparta} to explainable evaluation~\citep{zhao2022explainable}.

Recently, an emerging line of research is to exploit VLP models to upgrade OD models to solve the more challenging open-vocabulary setting, where a detector can generalize to new visual concepts with zero/few-shot adaption~\citep{gu2021open,kamath2021mdetr,li2022grounded,minderer2022simple}. Some of the VLP-based methods exploit large-scale visual grounding datasets for pretraining~\citep{kamath2021mdetr} and some of the work combines class-agnostic region proposal network (RPN) with a zero-shot image-text classifier respectively for localization and classification~\citep{zhong2022regionclip}. 

Unlike previous VLP-based methods that utilize one large vision-language corpus for pretraining, this paper explores a continual learning approach, i.e., \textit{can a detector learn from many OD datasets with increasing total visual vocabulary and eventually achieve the open-vocabulary detection capabilities?}. This approach is appealing for several reasons: (1) it opens the possibility of lifelong learning since one can improve a detector's zero/few-shot performance by feeding it with new datasets. (2) it is cost-effective since creating many small domain-specific datasets is much cheaper than creating a single large-vocabulary large dataset~\citep{gupta2019lvis}.  

On the other hand, joint training from multiple OD datasets with different labels faces two key technical challenges: \textbf{(1) taxonomy conflict:} each OD dataset is annotated with its pre-defined labels and a classic detector uses a fixed Softmax layer to classify object types~\citep{ren2015faster}. Such design forbids the possibility of learning from different label sets or dynamically adapting to new classes. \textbf{(2) fore/background inconsistency:} since the label set is different, then an object proposal may be considered as foreground in dataset A, while it is considered as background in dataset B. For example, an object ''cat'' is annotated in dataset A, but not in dataset B. Our study shows that this greatly hurts the multi-dataset performance of classic detectors since the RPN head is confused by the conflicting ground truth.

To address the above challenges, this work proposes a novel vision-language model, OmDet,  for open vocabulary object detection and phrase grounding. The main architecture novelty of OmDet is its latent query-centric fusion module that combines information from visual and text features and the proposed training mechanism that can easily accumulate knowledge from OD/grounding datasets from various domains. Two versions of OmDet is pre-trained, including OmDet V1 which is purely pre-trained on a large number of OD datasets (more than 100 domains), and OmDet V2 which is additionally pre-trained on visual grounding data~\citep{kamath2021mdetr}. 

The proposed method is evaluated on three downstream tasks: object detection in the wild (ODinW)~\citep{li2022elevater}, open-vocabulary detection, and phrase grounding~\citep{plummer2015flickr30k}. Results show that OmDet is able to outperform all prior art, including the powerful GLIP~\citep{li2022grounded} that is pre-trained on much larger datasets. Moreover, comprehensive model analysis is conducted to better understand the strength and limitations of OmDet. We conduct controlled study on joint training from four diverse datasets (COCO, Pascal VOC, and Wider Face/Pedestrian) and results show that our method is not only able to learn from all datasets without suffering from label and localization conflicts, but achieves stronger performance than single dataset detectors due to its share of knowledge among tasks. Also, we show that accumulating multiple datasets to expand to large vocabulary OD learning is an effective method to boost OmDet's zero/few-shot ability as well as parameter-efficient training performance (e.g., prompt tuning)

In summary, the contributions of this our paper are four folds:
\begin{itemize}
    \item We present OmDet, a novel language-aware OD architecture with Multimodal Detection Network (MDN) that can learn from any number of OD and grounding datasets. 
    \item Experiments show OmDet's state-of-the-art performance on well-known ODinW, open-vocabulary detection and phrase grounding benchmark.  
    \item Experiments confirm the effectiveness of the proposed multi-dataset training by solving the label difference and fore/background inconsistency challenges. 
    \item Experiments show that by scaling up visual vocabulary size via multi-dataset training, one can improve zero/few-shot and parameter-efficient fine-tuning.  
\end{itemize}
\color{black}

\section{Related Work}

\subsection{Vision-Language Pre-training}
One of the most studied topics of VLP is to pre-train massive image-text pair data. Recent advances in self-supervised learning have enabled models to learn rich representations from large-scale unlabeled data. 

For example, CLIP~\citep{radford2021clip} learns to predict which text matches which image, resulting in a versatile model that can perform well on various vision tasks without task-specific supervision. ALIGN~\citep{li2021align} further scales up CLIP by using a noisy dataset of over one billion image alt-text pairs. However, these models mainly focus on vision-based tasks and neglect the interaction between multiple modalities during pre-training. To address this limitation, several studies propose to learn joint multi-modal representations of image content and natural language for vision+language tasks (such as VQA and visual reasoning). Among them, OSCAR~\citep{li2020oscar}, UNITER~\citep{chen2020uniter} and VILLA~\citep{gan2020villa} adopt a two-stage approach: they first use an object detector (e.g., Faster R-CNN~\citep{zhang2021vinvl}) to extract vision features, then they apply a multi-layer transformer~\citep{vaswani2017attention} to the concatenation of the visual features and text features to learn joint embeddings.

Some studies propose to model visual input without relying on pre-trained object detectors. For instance, SOHO~\citep{huang2021seeing} uses a visual dictionary to extract compact image features from a whole image, which enables 10 times faster inference time than region-based methods. Similarly, ViLT~\citep{kim2021vilt} employs a vision transformer~\citep{dosovitskiy2020image} to capture long-range dependencies over a sequence of fixed-size non-overlapping image patches, without using convolutional visual features.

\subsection{Object Detection}
Objection detection, one of the predominant tasks in computer vision, aims to detect bounding boxes and classes of object instances. It has significantly evolved through the contributions of massive research in recent years. There are two major categories of detectors: two-stage and one-stage methods. Two-stage methods consist of a region proposal network (RPN) and a region-wise classifier. Classic models include R-CNN~\citep{girshick2014rich}, Fast R-CNN~\citep{girshick2015fast} and Faster R-CNN~\citep{ren2015faster}. One-stage methods eliminate the RPN stage and directly make final object predictions on the visual feature maps. Well-known systems include SSD~\citep{liu2016ssd}, Yolo~\citep{redmon2016you} and RetinaNet~\citep{lin2017focal}. Recently, end-to-end detectors such as DETR~\citep{carion2020end} have proposed to formulate the object detection task as a set prediction task. However, objection detection is often formulated as a closed-set problem with fixed and predefined classes and cannot handle object detection in the wild. To overcome the closed-set limitation, more realistic scenarios such as Multi-Dataset Object Detection (MDOD) and Open-Vocabulary Object Detection (OVOD) have attracted lots of attention.

\textbf{Multi-Dataset Object Detection:} MDOD focuses on increasing detectable object classes by training a single detector using multiple datasets. Traditional closed-set object detection demands training detectors on datasets with full annotations, and adding a new dataset means costly extra human annotations. Research on MDOD attempts to bypass the closed-set limitation, where a single detector is able to incrementally add object classes by adding new datasets with new classes. Yao et al.,~\citep{yao2020cross} proposes an MDOD framework with a preprocessed hybrid dataset and a dataset-aware focal loss. \citep{zhao2020towards} designs a conflict-free loss to avoid the ambiguity between positive and negative samples. Detection Hub\citep{meng2022detection} unifies multiple datasets with a query-based object detector with natural language embedding.

\textbf{Open-Vocabulary Object Detection:} OVOD, a more ambitious goal beyond the closed-set problem, refers to the capability of only training on annotated datasets and generalizing to unseen novel classes. Recently, OVOD has made such progress with the utilization of multi-modal vision-language pre-trained models~\citep{li2022grounded}\citep{zhou2022detecting}\citep{kamath2021mdetr}. RegionCLIP\citep{zhong2022regionclip} generates pseudo-labels for region-text pairs from caption datasets to perform regional vision-language pre-training and transfer to OVOD. ViLD\citep{gu2021open} proposed a two-stage open-vocabulary detector, which distills embeddings from teacher model CLIP ~\citep{radford2021learning} or ALIGN~\citep{jia2021scaling}. With inspiration from CoOp~\citep{zhou2022learning}, DetPro~\citep{du2022learning} introduces a technique to learn continuous prompt embedding that improves the performance of ViLD. OWL-ViT~\citep{minderer2022simple} transfers the pre-trained image-text model to the object detection by adding downstream detection heads and fine-tuning on OD datasets. 

\textbf{Object Detection as Grounding:} Phrase grounding refers to the process of identifying the relationship between individual phrases within a sentence and specific objects or regions depicted in an image ~\citep{kamath2021mdetr,deng2021transvg}. GLIP~\citep{li2022grounded} proposed that object detection can be viewed as a special case of phrase grounding. The authors of GLIP concatenate object types as a single string and ask the model to ground objects to word spans. This setup enables unified modeling between phrase grounding and object detection, and the resulting system achieves strong performance in long-tail object detection and zero-shot detection. 

Unlike previous grounding-based methods, the proposed method is designed to learn from an arbitrary number of object detection (OD) datasets, which does not necessarily need to train on grounding data. This ability is valuable for real-world scenarios, e.g., creating a multi-task OD model that simultaneously learns from many independent OD datasets.

\section{Our Approach}
\begin{figure*}[ht!]
\centering
  \includegraphics[width=14cm]{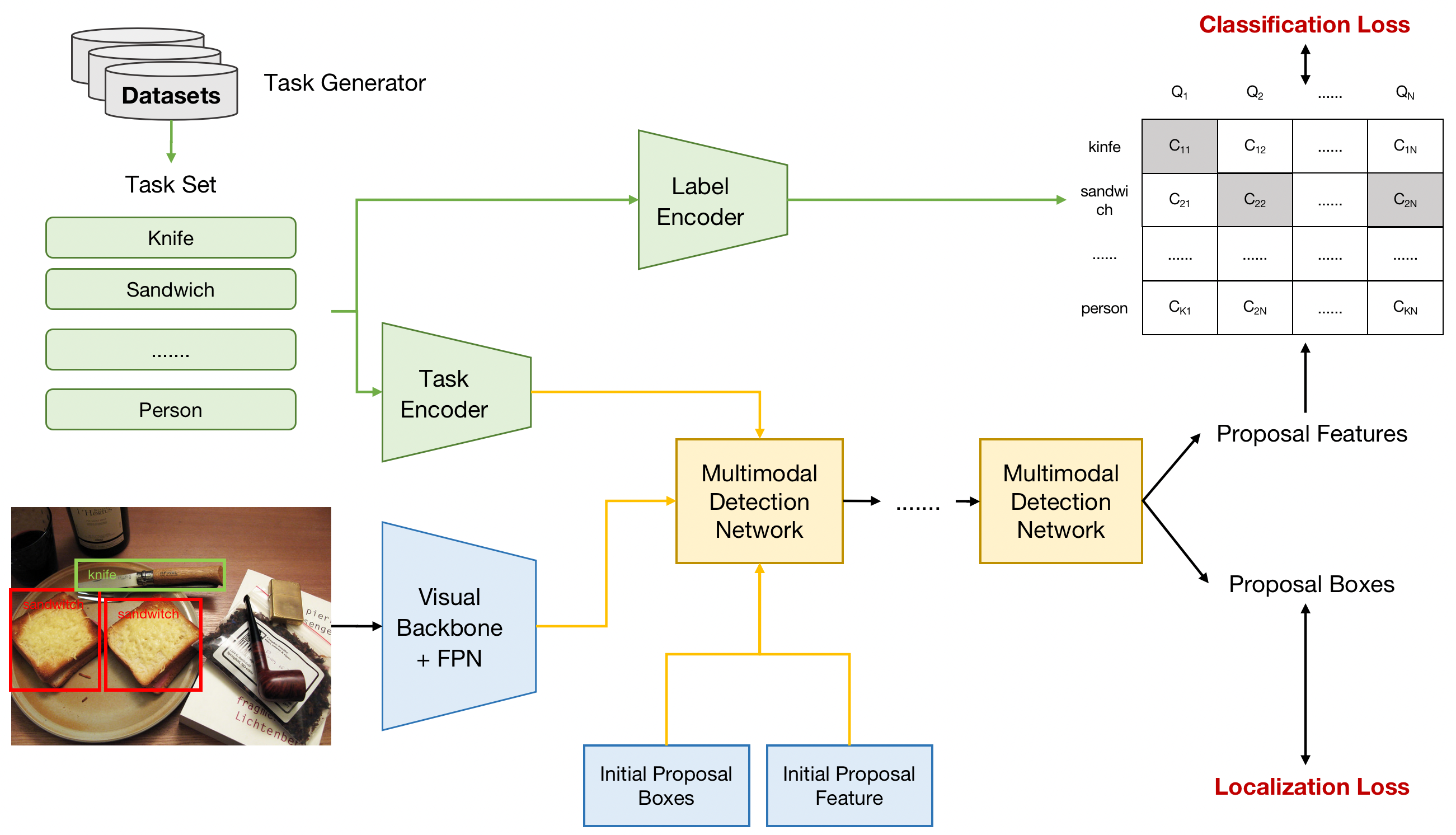}
  \caption{Overview of OmDet Architecture. The proposed Multimodal Detection Network iteratively fuses vision and language features into latent queries for object detection.}
  \label{fig:outline}
\end{figure*}

Before getting into the details of the proposed system, we first define the problem formulation. OmDet is designed for language-conditioned detection. Let $V$ be a large vocabulary of object types that OmDet can potentially detect. A task $T = \{w_1, w_2,... w_k\}$ is a set of $k$ object types that the model should detect in its forward path, where $w \in V$. Note that the size of $T$ can be dynamic ranging from 1 to $K$, where $K$ is the maximum supported number of object types in a single inference run. For the visual grounding setting, $T$ is the query sentence that contains $K$ word tokens. Meanwhile, Let $L$ be a set of natural language labels. In the object detection case, $L = T$. For the grounding cases, $L$ is the set of entities that appeared in caption $T$. Then given an input image $x$, a task $T$, and a label set $L$, the model is expected to detect all objects mentioned in $T$ from $x$. Since $T$ and $L$ are not fixed, an ideal model can dynamically adapt its detection targets conditioned on the task.

\subsection{Model Architecture}
\label{sec:3.1}
Following the above design principle, OmDet is introduced, a task-conditioned detection network that can learn from infinite combinations of tasks. It is composed of a vision backbone, a task encoder, a label encoder, and a multimodal detection network. The overall structure is illustrated in Fig\ref{fig:outline}. The following will describe each component in detail. 

\textbf{Vision Backbone}
Starting from the initial image $x_{img} \in R^{3 \times H_0 \times W_0}$ (with 3 color channels), let the vision encoder $f_v$ be a conventional Convolutional Neural Network (CNN)~\citep{liu2022convnet} or Vision Transformer (e.g. Swin Transformer~\citep{liu2021swin}). The vision encoder generates a lower-resolution visual feature map $f \in R ^{C \times H \times W}$ at each output layer. Then Feature Pyramid Network (FPN)~\citep{lin2017feature} is used to aggregate information from top to bottom and output a set of visual feature maps $\{P2, P3, P4, P5\}$. 

\textbf{Task Encoder and Label Encoder}
The term "task" refers to a natural language query designed to expand various text-aware vision tasks; (e.g., "Detect objects: \{\textit{the specified list of objects that we aim to identify}\}") The term 'label' refers to the language phrase output that is intended for detection purposes. The task set $T=\{w_1, w_2, ...w_k\} \in R^{k \times V}$ is set of natural language words. Then a task encoder $f_t$ or a label encoder $f_l$ is a transformer model that encodes the task set $T$ as a natural language sentence, and outputs a set of contextual word embeddings, i.e. $T_0=\{t_1, t_2, ...t_k\} = f_t(w_1, w_2, ...w_k) \in R^{k \times d}$ and $L=\{l_1, l_2, ...l_k\} = f_l(w_1, w_2, ...w_k) \in R^{k \times d}$, where $d$ is the contextual word embedding dimension size. We use pre-trained transformer-based language models, e.g. CLIP~\citep{radford2021clip} to initialize the task and label encoders. 

\begin{figure}[ht!]
\centering
  \includegraphics[width=0.75\textwidth]{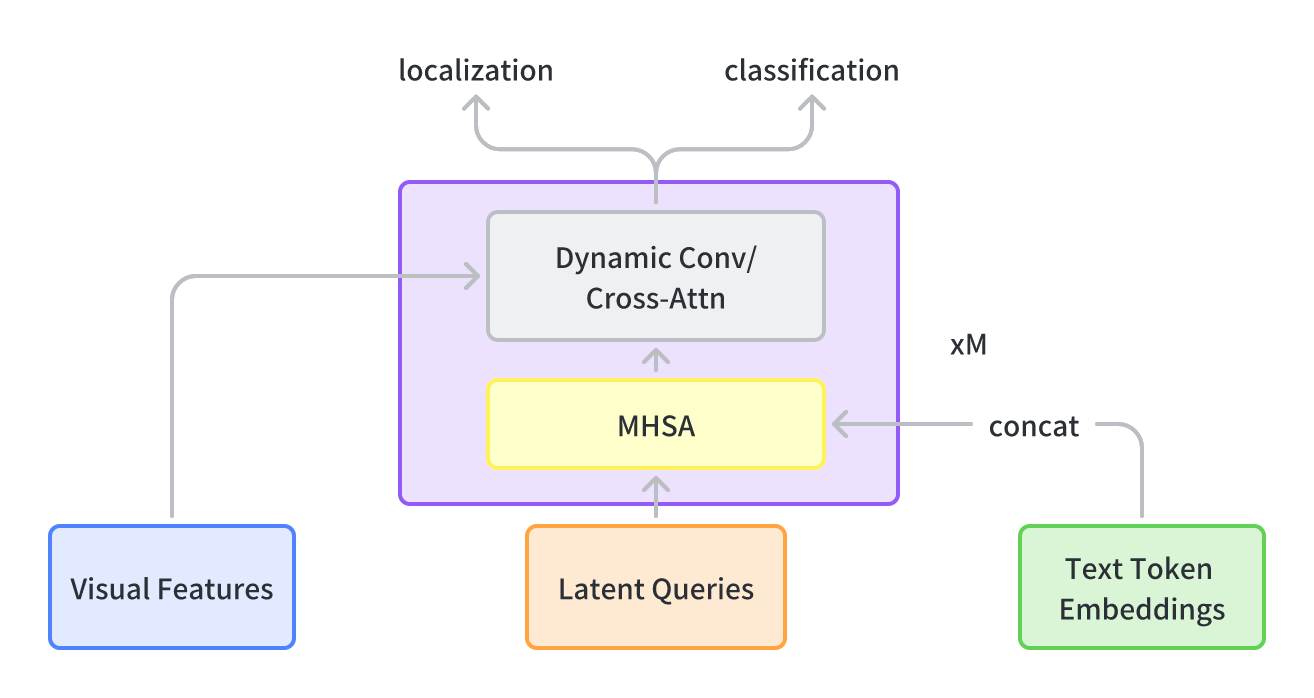}
  \caption{Network architecture for the Multimodal Detection Network (MDN), simplified here for illustration purposes. }
  \label{fig:mdn}
\end{figure}

\begin{figure*}[ht!]
\centering
  \includegraphics[width=14cm]{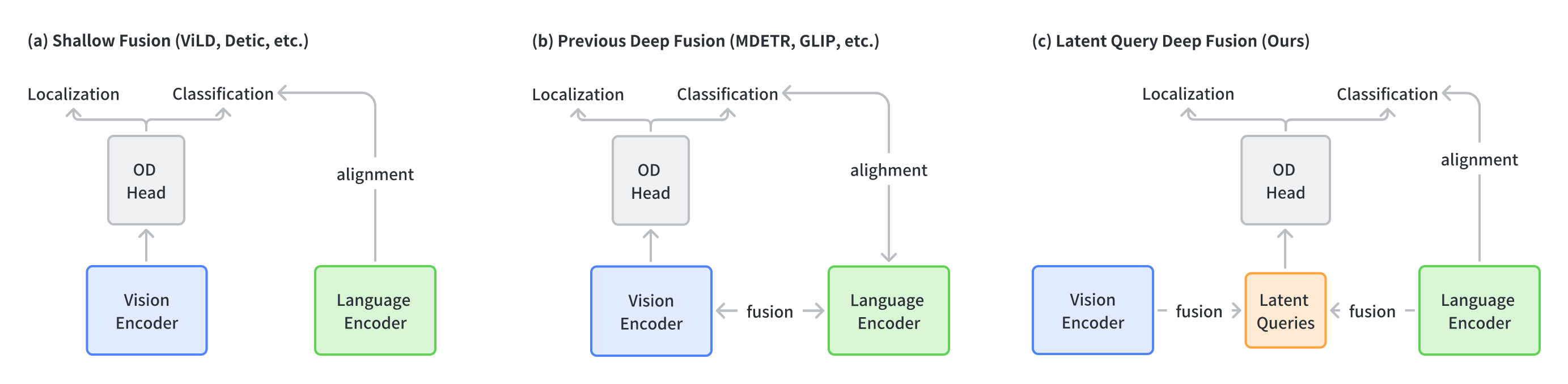}
  \caption{Comparison with other frameworks. (a) \textbf{Shallow fusion} that only utilizes text information for object classification.
  (b) \textbf{Deep fusion} that fuses visual and text in the backbone before entering the object detection head.
  (c) \textbf{Deep latent fusion (ours)} utilizes latent queries to fuse multimodal information, enabling adaption to any query-based OD architecture.}
  \label{fig:diff}
\end{figure*}

\textbf{Multimodal Detection Network}
The Multimodal Detection Network (MDN) is a core component of OmDet. Different from early work only fuse language and vision information in late stage~\cite{gu2021open}, we deploy deep fusion to combine information from the image and current task early on, in order to achieve strong performance. We are inspired by the Sparse-RCNN~\citep{sun2021sparse} network design and developed an iterative query-based fusion mechanism that fuses text features and visual features into latent queries. Figure~\ref{fig:diff} illustrates the differences between our method versus prior art. 
Let $Q \in R^{N \times d}$ be a fixed small set of learnable proposal features. The $N$ denotes the number of proposal features.  It is a set of high-dimensional (e.g., $d=256$) latent features that capture the rich information of a potential instance, by combining data from the vision backbone and contextual task embedding from the task encoder. Also, let $B \in R^{N \times 4}$ be a set of learnable one-to-one proposal boxes assigned to each feature. Then given the FPN output and task/label encoder output, the initial MDN operates as the following:
\begin{align}
v_0 & = \text{RoiPooler}(\{P2, P3, P4, P5\}, B_0)  \\
[Q_1, T_1] & = \text{MHSA}([Q_0, T_0]) \\
Q_2 & = \text{DynamicConv}(Q_1, v_0) \\
B_1 & = \text{RegHead}(Q_2) \\
C_1 & = \gamma cosine(\text{ClsHead}(Q_2), L) 
\end{align}
where $T_i$ is the task embedding at iteration $i$ and $L$ is the label embedding. Note that MDN can be stacked to iterative refine its output the same as Sparse-RCNN, with the key difference that $T_i$ is fused with the proposal feature before the Dynamic Convolution layer and also $T_i$ is also iteratively updated at each run of MDN block. This enables the network to learn to adjust the task embedding and the proposal embedding jointly and adapt both object localization and classification heads conditioned on the given task. Figure~\ref{fig:mdn} shows the process by which MDN first combines information between latent queries and language embedding via MHSA, and then infuses visual features with DynamicConv.  Note that we can easily adapt MDN to other query-based detectors such as DETR~\cite{carion2020end}, in which the DynamicConv operation is replaced by a CrossAttention module.

With the utilization of deep fusion between image features and task embedding at MDN, the challenge of fore/background inconsistency is solved. Other models like \citep{zhou2022detecting}\citep{minderer2022simple} try to solve the fore/background inconsistency by training a perfect RPN to find all possible objects, which is hard to achieve. Our method applies deep fusion at an early stage to help the model be conscious of fore/background according to task embedding, and therefore properly switching fore/background among different tasks. To handle the taxonomy conflict, the label encoder is applied to get the text embedding of the target label, then the label embedding is passed to the classification stage to eliminate naming differences. Taxonomy conflict is solved by projecting the target label into embedding space since the same object with different naming will be close to each other.

\subsection{Model Training}
\textbf{Set Prediction Loss} Given the proposed model, it uses set prediction loss~\citep{carion2020end} on the fixed-size set of predictions of classification and box coordinates. Set-based loss produces an optimal bipartite matching between predictions and ground truth objects using the Hungarian algorithm. The matching cost is defined as follows:
\begin{equation}
    L = \lambda_{cls} \cdot L_{cls} + \lambda_{L_1} \cdot L_{L_1} + \lambda_{giou} \cdot L_{giou}
\end{equation}

Here $L_{cls}$ is focal loss~\citep{lin2017focal} of predicted classifications and ground truth category labels, $L_{L_1}$ and $L_{giou}$ are L1 loss and generalized IoU loss~\citep{carion2020end} between normalized center coordinates and height and width of predicted boxes and ground truth box, respectively. $\lambda_{cls}$, $\lambda_{L_1}$ and $\lambda_{giou}$ are coefficients of each component. The training loss is the same as the matching cost except that only performed on matched pairs. The final loss is the sum of all pairs normalized by the number of objects inside the training batch.

\textbf{Task-Sampling Strategy} 
For \textit{object detection datasets}, in order to simulate a diverse set of tasks for meta-learning during training and also enforce the model to condition its output on a given task, a novel task sampling strategy is used during training. 
\begin{enumerate}
    \item Let the max size of a given task be $K$, for an image $x$ from a dataset $d$ in the mini-batch, we first sample $k \in [1, K]$ with a uniform distribution. 
    \item Let the number of unique object types in $x$ be $m$, if $m > k$, then only a random subset of $k$ object types are kept and the extra annotations are removed for this mini-batch. If $m < k$, then additional negative object types are randomly selected from the vocabulary $V$ of dataset $d$. 
    \item The model is trained with the above-sampled task and ground truth annotations.
\end{enumerate}

With the above method, each image in every mini-batch will have a different set of tasks to learn from. When we learn from a large-vocabulary object detection dataset, e.g., LVIS, which contains 1200 unique object types, the unique combination of task size $k$ is $C_k(n) = \frac{1200!}{(1200-k)!k!}$. If $k=20$, then it produces 1.34E43 possibilities, a quite large number. Experiments show that the proposed training strategy serves the purpose well, and yields models that perform task-conditioned object detection.

For learning from \textit{phrase grounding dataset}, the task $T$ is simply the corresponding caption of the image. The label set $L$ is the set of entities that appeared in the caption. However, since there are only a few entities in each caption, learning of $L_{cls}$ becomes too easy. Therefore, we randomly select from other entities in the dataset to create a label set up to $K$ classes to increase the difficulty of learning. This method is proven to be effective in improving performance on phrase grounding in later experiments. 

\subsection{Comparison to Grounding-based Method}
Our proposed architecture, the Multimodal Detection Network, has several strengths over traditional approaches that directly fuse text and vision features. Instead, our model fuses latent queries with text features, leading to the following advantages:

\textit{Deep fusion for any query-based OD:} early VLP work, e.g., ViLD~\citep{gu2021open} and Detic~\citep{zhou2022detecting}, use shallow fusion for object detection, i.e. use text embedding only for classification, which cannot solve fore/background conflicts. Meanwhile, prior deep fusion models, e.g., MDETR~\citep{kamath2021mdetr} and GLIP~\citep{li2022grounded}), use specialized cross-attention architecture to fuse the text and visual features. Our method can be applied to any query-based OD architecture, e.g. DETR, Sparse-RCNN, without the need for model change.  

\textit{Inference speed and performance}: visual grounding MDETR~\citep{kamath2021mdetr} and TransVG~\citep{deng2021transvg} models encode one class at a time for OD and suffer from slow inferences speed, e.g. 10s/image for MDETR. Also, MDETR uses a transformer to fuse images with text, which cannot scale up to multi-scale features due to the complexity of self-attention. Our method deals with fixed-size latent queries, which are independent of visual features. Thus, our method is able to predict many classes with significant speed up with on-par or better performance.

\section{Experiments}
\subsection{Implementation Details}
We implement OmDet with the following settings:

For text embeddings,  CLIP-B/16 text encoder~\citep{radford2021learning} is used throughout the study. We did not use the prompt template as used in study~\citep{gu2021open}, i.e. encoding object names in a template \textit{a photo of \{\}}. This is because preliminary studies show no major difference between using versus not using the prompt template. Furthermore, the preliminary study also suggests there are no significant differences between using single-modal language models, e.g. BERT~\citep{devlin2018bert} and RoBERTa~\citep{liu2019roberta}, versus multimodal-language models e.g. CLIP. We suspect this is because object detection does not involve complex language understanding. 

The task and label encoders share the same text encoders. On top of the text encoder, two independent Transformers layers~\citep {vaswani2017attention} are used to further dedicated encoding for task input and label input. Study shows that the set encoding is able to improve OmDet's performance. 

For visual backbones, both Swin Transformers~\citep{liu2021swin} and ConvNeXt~\citep{liu2022convnet} are used in the experiments. A standard FPN~\citep{lin2017feature} is used to extract a four-level feature map form the visual encoders. Both backbones are pre-trained on ImageNet 21K data~\citep{ridnik2021imagenet}. Preliminary studies found that ConvNeXt usually performs on par or better than Swin Transformers. Therefore, we use ConvNeXt as the default choice. 

Lastly, the MDN network utilizes MHSA to fuse information from visual input and text input to latent queries. We equip MDN with 300 latent queries and we use ROIAlignV2~\citep{he2017mask} as the ROI Pooler to extract region features from the visual backbone. 6 sequential MDN blocks are cascaded to create the final bounding boxes and classification prediction. 

\subsection{Large-scale Pre-training}
Two versions of large-scale pre-training are conducted. 

\textbf{Large-scale OD Pre-training (OmDet V1)}: in this setting, we accumulate a large number (104) of object detection datasets for pre-training to show that OmDet is able to accumulate knowledge from many OD datasets without suffering from fore/background and label inconsistency challenges. Pre-training datasets include COCO~\citep{lin2014microsoft}, Object365~\citep{shao2019objects365}, LVIS~\citep{gupta2019lvis}, PhraseCut~\citep{wu2020phrasecut} and Roboflow 100~\citep{ciaglia2022roboflow}. Data details are described in Table~\ref{tbl:od_pretrain}
\begin{table}[ht!]
\centering
\begin{tabular}{l|lll} \hline
          & \# Classes & \# Images & Federated\\ \hline
COCO      & 80         & 100K      & No \\
O365      & 365        & 2M        & No\\
LVIS      & 1203       & 100K      & Yes \\
PhraseCut & 3013       & 70K       & Yes \\
RF100      & 829        & 224K      & Yes \\ \hline
\end{tabular}
\caption{Pre-train data used in large-scale OD pre-training, resulting in OmDetV1.}
\label{tbl:od_pretrain}
\end{table}

\textbf{Large-scale OD \& Grounding Pre-training (OmDet V2)}: In the second version, we exclude any images related to COCO and LVIS datasets from pre-training since we will test zero-shot performance on these two datasets. In addition to large-scale OD multi-dataset pre-training, OmDet is able to horizontally expand to the non-OD type of training data. Specifically, we include the GoldG grounding dataset curated by~\cite{kamath2021mdetr}, which includes 1.3M pairs of image-caption data with grounded entities. Data details are described in Table~\ref{tbl:ground_pretrain}

\begin{table}[ht!]
\centering
\begin{tabular}{l|lll} \hline
          & \# Classes & \# Images & Type\\ \hline
O365      & 365        & 2M        & OD\\
PhraseCut & 3013       & 70K       & OD \\
RF100      & 829        & 224K      & OD \\
GoldG      & 1.3M      & 100K      & Ground \\ \hline
\end{tabular}
\caption{Pre-train data used in large-scale OD \& Ground pre-training, resulting in OmDetV2.}
\label{tbl:ground_pretrain}
\end{table}

\begin{table*}[!h]
\centering
\small
\begin{tabular}{l|llll}
     	\hline
        Models & Fusion & Backbone & \#Param &  Pretrain Data \\
         \hline
         DyHead-T & - & Swin-T & 30M & - \\ 
         DINO-T & - & Swin-T & 50M & O365 \\\hline
         MDETR & Deep & RN101 & 185M & GoldG \\
         Detic-T & Shallow & ConvX-T & 138M & COCO,LVIS,IN-21K  \\
         GLIP-T & Deep & Swin-T  & 231M & O365,GoldG,Cap4M\\ \hline
         OmDetV1-T & Deep Latent & ConvX-T & 180M & \multirow{2}{2.7cm}{COCO,LVIS,O365,\\ PhraseCut, RF100} \\ \\
         OmDetV1-B  & Deep Latent & ConvX-B  & 240M \\ \hline
        OmDetV2-T & Deep Latent & ConvX-T & 180M & \multirow{2}{2.7cm}{O365, GoldG\\ PhraseCut, RF100} \\ \\
         OmDetV2-B  & Deep Latent & ConvX-B  & 240M \\ \hline
\end{tabular}
\caption{Baseline models and their training setup.}
\label{tbl:setup}
\end{table*}

\textbf{Model Training:} For OmDet models, the initial learning rate is 5e-5 and it decays at 70$\%$ and 90$\%$ of total iteration steps by 0.1. ConvNeXt Base backbone is used with a 6-layer MDN head. The batch size is 40 and the maximum number of detections per image is 300 and K is set to 80. All of the proposed models are pre-trained for 36 epochs using NVIDIA A100 GPU cluster and then fine-tuned on the downstream data.

\subsection{Downstream Tasks}
We focus on three types of downstream tasks for evaluation:

\textbf{Object Detection in the Wild}: object detection in the wild test a model's ability to adapt to various different domains with drastically different label sets. ELEVATER~\citep{li2022elevater} is a new object detection benchmark that is composed of 35 diverse real-world challenging domains with full-shot, few-shot and zero-shot training settings. Note that there are two variations of data used in prior work~\citep{li2022grounded}. The full version includes 35 domains, which we will refer to as ODinW35, and the second version only includes 13 out of 35 domains, which we will refer to as ODinW13. The evaluation metric is AP.

\textbf{Open-vocabulary Object Detection}: open-vocabulary detection tests models' ability to recognize a large number of objects with types that are not included in the training. Zero-shot performance on COCO~\citep{lin2014microsoft}, LVIS~\citep{gupta2019lvis}, and ODinW~\citep{li2022elevater} are commonly used as the benchmark. The evaluation metric is AP. 

\textbf{Phrase Grounding}: For phrase grounding on Flickr30k~\citep{plummer2015flickr30k}, we do not further fine-tune the model after grounding pre-training, and just directly evaluate on the Recall@ 1,5,10 metrics.

We provide the detailed settings of the baseline models used in our experiments, including information on the fusion (deep vs. shallow), backbones, number of parameters and pretraining data (Table ~\ref{tbl:setup})

\section{Main Results}

\subsection{Results on Object Detection in the Wild}
In our evaluation of ODinW, we compared the zero-shot, few-shot, and full-shot scores of GLIP-Tiny, DyHead-Tiny, DINO Swin-Tiny, and OmDet (Table~\ref{tab:ELEVATER_Compare_others}). When compared to state-of-the-art models that were trained with the same backbone size, our OmDetV1-T model achieved the highest AP scores on few-shot and full-shot evaluations. Furthermore, we trained OmDet V1-T and OmDet V1-B under the same settings, but with different backbones. OmDet V1-B outperformed all other models and achieved state-of-the-art results on zero-shot, few-shot, and full-shot evaluations. Note that the word definition of Wiktionary is used as a knowledge source for OmDet v1-B on the zero-shot setting. 
\begin{table*}[ht]
\centering
\begin{tabular}{l | c c c c}
     	\hline
        Models & Zero-shot & Few-shot & Full-shot \\
         \hline\hline
         GLIP-T~\citep{li2022elevater} & 19.6 & 38.9 & 62.6  \\ 
         DyHead-T~\citep{dai2021dynamic} & - & 37.5 & 63.2  \\ 
         Detic-T~\citep{zhou2022detecting}  & 13.0 & - & 62.6 \\
         DINO Swin-T~\citep{zhang2022dino} & - & 41.2 & 66.7 \\
         \hline
         OmDetV1-T (ours) & 15.6 & 42.3 & 67.61 \\
         OmDetV1-B (ours) & \textbf{19.7} & \textbf{45.4} & \textbf{68.85} \\
         \hline
\end{tabular}
\caption{Comparison between OmDetV1 and other models, on average AP of zero-shot, few-shot (3-shot), and full-shot on ODinW35.}
\label{tab:ELEVATER_Compare_others}
\end{table*}

\subsection{Results on Open-Vocabulary Detection}
We evaluated the zero-shot performance of different models on several open-vocabulary detection datasets, including COCO Val, LVIS MiniVal, ODinW13, and ODinW35 (Table~\ref{tab:open_voc_det}). 
Our proposed models, OmDetV1-B and OmDetV2-B, achieved competitive results on the evaluated datasets. Specifically, OmDetV2-B significantly outperformed all other models on all datasets, achieving 9 points higher AP than the previous state-of-the-art model (GLIP-B) with the same base backbone on COCO Val.
Moreover, our model showed exceptional performance on rare objects in LVIS MiniVal, outperforming the previous state-of-the-art model OWL-B by almost 8 points (20.8 vs. 27.92). These results demonstrate the effectiveness of our proposed models for open-vocabulary detection tasks, particularly on rare object detection where data scarcity is an issue. Thus, our model achieved the highest performance on ODinW, which is a dataset that contains a large number of rare objects and reflects real-world applications.
\begin{table*}[ht!]
\centering
\begin{adjustbox}{width=\textwidth}
\begin{tabular}{llllllll} \hline
            & COCO Val & \multicolumn{4}{c}{LVIS MiniVal} & ODinW35 & ODinW13 \\ \hline
            & AP            & APr     & APc    & APf   & AP    & AP      & AP      \\ \hline \hline
MDETR~\citep{kamath2021mdetr}       & -             & -       & -      & -     & -     & 10.7    & 25.4    \\
OWL-B~\citep{minderer2022simple}     & 30.9          & 20.8    & -      & -     & 17.1  & 18.8    & 40.9    \\
GLIP-T~\citep{li2022grounded}      & 46.7          & 17.7    & 19.5   & 31.0  & 24.9  & 19.6    & 44.4    \\
GLIP-B~\citep{li2022grounded}      & 48.1          & 17.0    & 23.9   & 35.9  & 29.1  & -       & 44.8    \\ \hline
OmDetV1-B (ours)  & -             & -       & -      & -     & -     & 19.7    & 42.1    \\
OmDetV2-B (ours)   & \textbf{57.1} & \textbf{27.92}   & \textbf{32.6}   & \textbf{36.4}  & \textbf{34.1}  & \textbf{20.9}   & \textbf{47.2}   \\ \hline
\end{tabular}
\caption{Zero-shot Performance on COCO Val 2017, LVIS MiniVal, and ODinW 13/35 for open-vocabulary detection. In the terms APr, APc, and APf, the letters r, c, and f denote rare, common, and frequent, respectively.}
\label{tab:open_voc_det}
\end{adjustbox}
\end{table*}
\subsection{Results on Phrase Grounding}

\begin{table*}[ht!]
\centering
\begin{tabular}{llllll} \hline
           & Backbone   & \multicolumn{3}{l}{Flickr30K Val}   \\ \hline
           &                       & R@1           & R@5  & R@10         \\ \hline\hline
VisualBERT~\citep{li2019visualbert} & ResNet-101            & 70.4          & 84.5 & 86.3  \\
MDETR~\citep{kamath2021mdetr}      & EN-B5      & 83.6          & 93.4 & 95.1  \\
GLIP-B~\citep{li2022grounded}      & Swin-B     & \textbf{85.7}          & 95.0 & 96.2\\
OmDetV2-B  & ConvX-B     & 85.1          & \textbf{96.2} & \textbf{97.5} \\ \hline
\end{tabular}
\caption{Zero-shot Performance on Flickr30K val for Phrase Grounding. }
\label{tab:phrase_grounding}

\end{table*}

We compared our model's performance to that of VisualBERT, MDETR, and GLIP-B, which use ResNet-101, EfficientNet-B5, and Swin-B backbones, respectively. Table~\ref{tab:phrase_grounding} shows the results of the evaluation in terms of Recall@1, Recall@5, and Recall@10.

Our model achieved a competitive performance on this task, with Recall@1, Recall@5, and Recall@10 scores of 85.1, 96.2, and 97.5, respectively. Our model's performance is only slightly lower than that of the state-of-the-art model, GLIP-B (85.7 vs. 85.1) at Recall@1, this is due to our approach of fixed-size latent queries independently over contexts to reduce complexity. However, we achieved higher scores than GLIP-B at Recall@5 and Recall@10 with significantly more efficient computations than the traditional approaches. 

These results demonstrate the effectiveness of our OmDet for the task of phrase grounding as well and highlight the importance of incorporating latent query deep fusion in both object detection and phrase grounding.

\section{Ablation and Analysis}
To further investigate the behavior of our proposed method, we conducted several follow-up studies: 1) an analysis of the efficacy of deep fusion, 2) an analysis of the effect of pre-training, and 3) visualizations of language-aware object detection. 
\subsection{Analyze the Efficacy of Deep Fusion}
\begin{table*}[!ht]
  \begin{adjustbox}{width=\textwidth}

\centering
\scalebox{1.0}{
\begin{tabular}{ll|ll|ll|ll|ll}\hline
 \multicolumn{1}{c}{}& \multicolumn{1}{c}{Fusion}& \multicolumn{2}{c}{COCO}&  \multicolumn{2}{c}{PASCAL VOC}& \multicolumn{2}{c}{WIDER FACE} & \multicolumn{2}{c} {WIDER Pedestrain} \\ 

 & - & AP & AP50  & AP & AP50  & AP & AP50  & AP & AP50 \\ \hline \hline
 
 Sparse R-CNN & N/A & 41.76 & 61.07  & 44.35 & 64.8 & 22.69 & 42.71 & 35.03 & 56.59 \\ 
 OmDet-Single & N/A & \textbf{43.03} & \textbf{62.44} & 45.53 & 65.85 & 23.39 & 44.82  & 33.93 & 55.16  \\ \hline
 OmDet-Shallow & Shallow & 35.32 & 50.64  & 55.41 & 73.94  & 22.94 & 43.48  & 34.97 & 55.09 \\ 
\textbf{OmDet} & Deep & 42.15 & 61.39  & \textbf{60.85} & \textbf{82.13} & \textbf{30.72} & \textbf{57.18}  & \textbf{46.78} & \textbf{73.99}  \\ \hline

\end{tabular}}
  \end{adjustbox}

\caption{MDOD training results on four datasets. OmDet is able to resolve task conflict issues in MDOD and achieves higher performance compared to single dataset models.}
\label{tab:four-data}

\end{table*}

\begin{figure*}[ht!]
\centering
  \includegraphics[width=14cm]{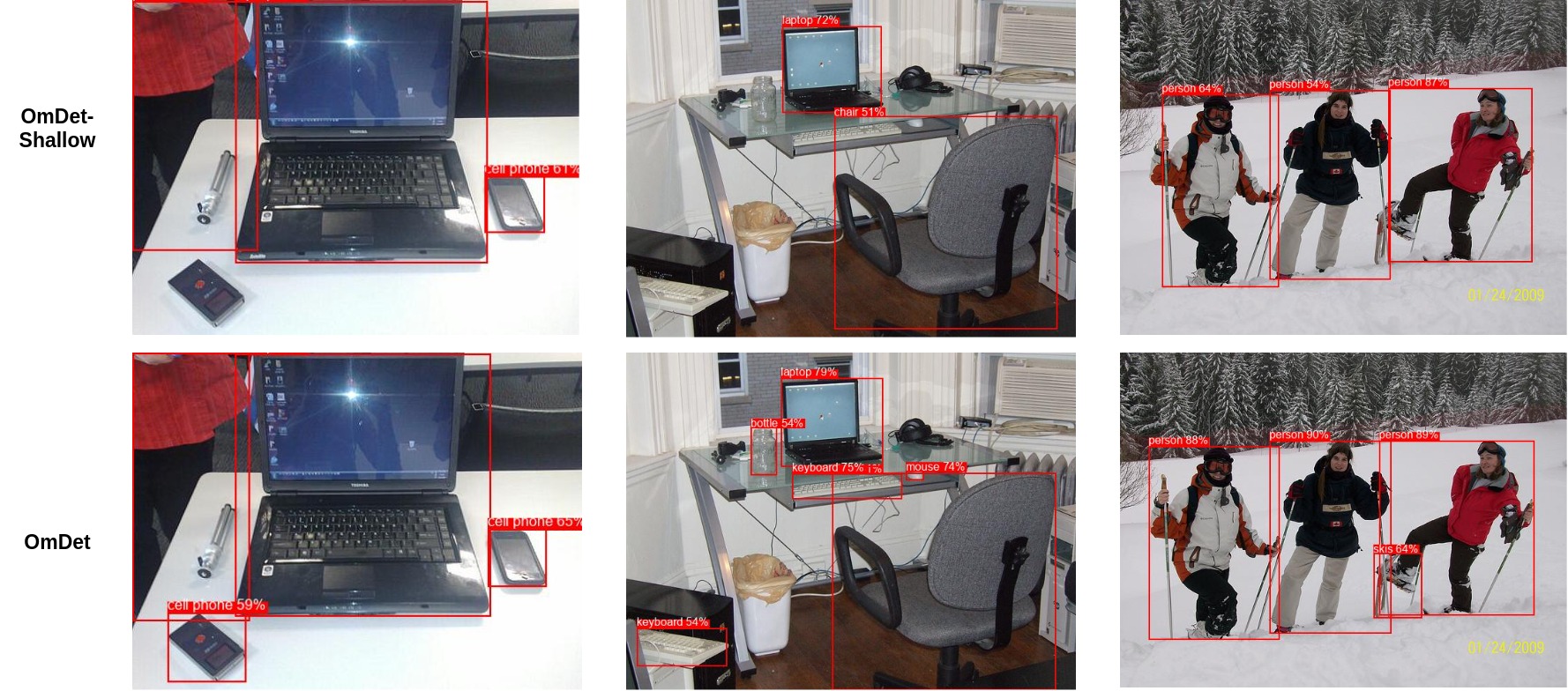}
  \caption{Inference results on COCO, showing that OmDet-shallow suffers from fore/background inconsistency (missing objects), while OmDet does not.}
  \label{fig:Shallow vs Deep}
\end{figure*}
We verify that the proposed deep fusion mechanism can effectively learn from multiple object detection datasets without suffering from the task conflict challenge. Additionally, we show the benefit of the proposed multiple datasets training over a single dataset.

For MDOD, we follow the experimental setting from \citep{yao2020cross} and choose COCO\citep{lin2014microsoft}, Pascal VOC\citep{everingham2010pascal}, WIDER FACE\citep{yang2016wider} and WIDER Pedestrian\citep{loy2019wider} as joint-training datasets.

Note that COCO is the larger data with 118K images while the other 3 datasets are almost 10 times smaller. Also the COCO dataset has a diverse set of categories that cover the classes in Pascal VOC and WIDER Pedestrian. WIDER Face is the only dataset that has ''face'' class. Therefore, these four datasets serve as a great testing bed for MDOD study.
Three baselines are served as baselines. First, we include Sparse R-CNN~\citep{sun2021sparse} as the baseline due to its strong performance and similarity to OmDet in terms of model structure. Then we create OmDet-{Single}, OmDet-{Shallow} for the ablation study. 

\begin{itemize}
    \item \textbf{OmDet-Single}: To compare the performance on single datasets with Sparse R-CNN, we train OmDet on the four datasets separately. Since they are only trained with a single dataset, so it cannot benefit from the proposed multi-dataset training. 
    \item \textbf{OmDet-Shallow}: We also train an OmDet-Shallow model by removing the task encoder and MDN, which degenerates the localization network to Sparse R-CNN, and only utilize the language feature for the final label classification. This is similar to previous work such as Detic\citep{zhou2022detecting}.
\end{itemize}

We use Image Net\citep{deng2009imagenet} pre-trained Swin Transformer Tiny\citep{liu2021swin} as the visual backbone and use CLIP ViT-B/16 as language encoder for OmDet. The same Swin Transformer is used as the backbone for Sparse R-CNN. All models are trained with 12 epochs. The initial learning rate is set to 5e-5 for OmDet and 2.5e-5 for Sparse R-CNN. 

\textbf{OmDet vs. Sparse R-CNN on single Dataset}:
First, we demonstrate the validity of our framework on classic OD tasks. 

As shown in Table \ref{tab:four-data}, OmDet-Single gets higher AP scores on COCO, PASCAL VOC and WIDER FACE than Sparse R-CNN under the same training setting, and is only about 1 point lower on WIDER Pedestrian. These results prove our model with the novel language-aware OD architecture still maintains the same or better performance on single dataset OD using the same number of trainable parameters, but since it is only trained with a small set of vocabulary, it does not have the open-vocabulary or few-shot capability.

\textbf{OmDet vs. OmDet-Single}:
The only difference between OmDet and OmDet-Single is that OmDet is joint-trained on all four datasets by utilizing the proposed language-aware OD architecture. 
Table~\ref{tab:four-data} shows that the AP scores of OmDet are significantly higher than OmDet-Single on PASCAL VOC (+15.52 AP), WIDER FACE (+7.33 AP), and WIDER Pedestrian (+12.85 AP). Moreover, OmDet shows better performances than Sparse R-CNN on all datasets. These results confirm that OmDet possesses the capability of multi-dataset training by solving taxonomy conflicts and fore/background inconsistency. Moreover, knowledge sharing in joint training improves the overall detection performance, especially for the ones with fewer training samples.

\textbf{OmDet vs. OmDet-Shallow:} Lastly an ablation study is used to verify the contribution of the proposed MDN block.
OmDet's fusion mechanism is \textit{deep} since the task embedding is combined with visual features early on and influences both localization and classification.  
On the other hand, OmDet-Shallow's fusion mechanism is \textit{shallow}, i.e., it only utilizes the label embedding in the final layer of object classification. 

Table~\ref{tab:four-data} shows that OmDet performs stronger and OmDet-Shallow only partially solve the MDOD challenges. OmDet-Shallow achieves good performance on PASCAL VOC, WIDER FACE, and WIDER Pedestrian compared with OmDet-Single. This is because OmDet-Shallow resolves the taxonomy conflict challenge and enables semantic sharing among object label embeddings, similar to Detic~\citep{zhou2022detecting}.

However, OmDet-Shallow fails on COCO with a low AP score, since it cannot resolve the fore/background inconsistency challenge. Since COCO has 80 categories, which is much larger than the other three datasets, many of its objects are considered as background in the other three datasets. Therefore, the low AP is caused by incorrectly detecting COCO objects as background.
We visualize outputs of OmDet-Shallow and OmDet on COCO images in Figure \ref{fig:Shallow vs Deep}, which confirms our hypothesis that OmDet-Shallow detects many objects that are not in Pascal VOC and Wider Face/Pedestrian as background.
More examples can be found that although OmDet-Shallow correctly detects all pedestrians in the last images, it misses object "Skis". Unlike OmDet-Shallow, OmDet has benefited from deep fusion and detects all the images correctly.

\begin{table*}[ht]
\centering
  \begin{adjustbox}{width=\textwidth}

\begin{tabular}{l | c| c| c c c}
     	\hline
        Models & Pre-train Data & Zero-shot & Full-model FT & Head-only FT & Prompt FT \\
         \hline\hline
         OmDet-C & C & 9.8 & 61.7 & 54.7~\color{gray}{(-11.3\%)} & 21.0~\color{gray}{(-65.9\%)} \\
         OmDet-CO & C,O & 13.5 & 63.2 & 56.8~\color{gray}{(-10.1\%)} & 24.8~\color{gray}{(-60.7\%)} \\
         OmDet-COL & C,O,L & 12.4 & 63.2 & 57.6~\color{gray}{(-8.8\%)} & 25.5~\color{gray}{(-59.6\%)} \\
         OmDet-COLP & C,O,L,P & 13.5 & 63.0 & 58.5~\color{gray}{(-7.1\%)} & 29.3~\color{gray}{(-53.4\%)} \\
         OmDet-COLPG & C,O,L,P,G &\textbf{15.6} & \textbf{65.2} & \textbf{59.8}~\color{gray}{(-6.1\%)} & \textbf{34.7}~\color{gray}{(-45.5\%)} \\
         \hline
\end{tabular}
  \end{adjustbox}

\caption{Average AP of zero-shot, full-model, head-only and prompt fine-tuning on 35 downstream tasks in ODinW. The gray text shows the performance drop of parameter-efficient tuning compared to full-model tuning. All models here use ConvNeXt Tiny backbone.(C: COCO, O: Obj365, L: LVIS, P: Phrasecut, and G: GCC3M)}
\label{tab:ELEVATER}
\end{table*}
\subsection{Analyze the Effect of Pre-training}
To investigate the impact of pretraining settings, we conducted a series of experiments by gradually increasing the number of pretraining datasets. Specifically, we added 5 intermediate variations based on Table~\ref{tab:ELEVATER}. The aim of this setup is to examine the relationship between the number of visual concepts in the pretraining data and the performance of the model on downstream tasks under various fine-tuning settings. Note that we used OmDet ConvNeXt-T as the backbone architecture for our ablation studies. 

\textbf{The effectiveness of Zero/Few-Shot :} As shown in Table~\ref{tab:ELEVATER}, adding more pre-train datasets yields significant improvement in zero-shot settings. Specifically, adding the object365 dataset gives an absolute gain of 3.7 points on the average mAP. Surprisingly, adding LVIS to the pre-train data hurts performance by 1.1 points. We speculate that the performance drop is due to the noisy and incomplete annotations of LVIS dataset. Adding GCC dataset to the pre-train corpora yields another huge gain, leading the zero-shot performance to 16.0 (compared to 9.8 for OmDet-C). There are several promising directions to further improve the zero-shot performance OmDet, including unfreezing the text encoder during pre-training and incorporating phrase grounding data with contextual text information in pre-training. We leave them to future research.

Meanwhile, the 35 downstream tasks in ODinW come with different training data sizes, varying from only 17 training images to more than 32K training images. Therefore, we divide the 35 tasks into three categories: (1) Small-shot (8 tasks): tasks with less than 200 training images (2) Medium-shot (13 tasks): tasks with between 200 to 2000 training data (3) Big-shot (14 tasks): tasks with more than 2000 training images. Results with full-model fine-tuning are summarized in Table~\ref{tab:fewshot}. Results show that large-scale multi-dataset pre-training is particularly effective for small-shot and medium-shot tasks with limited in-domain training data. Especially for small-shot datasets, OmDet outperforms OmDet-C with 10.99 absolute AP points. Whereas for Big-shot tasks, the advantages of pre-training become less evident.

\begin{table}[!ht]
\centering
\begin{tabular}{l|c c c}
\hline
        Models & Small-Shot & Medium-Shot & Big-Shot \\
         \hline\hline
         OmDet-C & 49.48 & 57.09 & 70.16  \\
         OmDet-CO & 54.37 & 58.89 & 70.98  \\
         OmDet-COL & 55.07 & 57.99 & 71.22  \\
         OmDet-COLP & 53.44 & 58.05 & 70.94  \\
         OmDet-COLPG & \textbf{60.47} & \textbf{61.25} & \textbf{71.65} \\
         \hline
\end{tabular}
\caption{Average AP of full-model fine-tuning on 35 downstream tasks in ODinW for Small-shot, Medium-Shot and Big-Shot tasks.}
\label{tab:fewshot}
\end{table}

\textbf{Parameter-efficient Fine-tuning: } As large-scale pretraining models get significantly larger, e.g., more than 1B parameters, the cost to fine-tune (FT) the entire model becomes prohibitive for low-end GPUs. Parameter-efficient fine-tuning is designed to alleviate this challenge by only tuning a very small proportion of the entire model. In this paper, we explore two options: Head-only Tuning and Prompt Tuning. 

\begin{figure} 
\centering 
\includegraphics[width=0.45\textwidth, height=4cm]{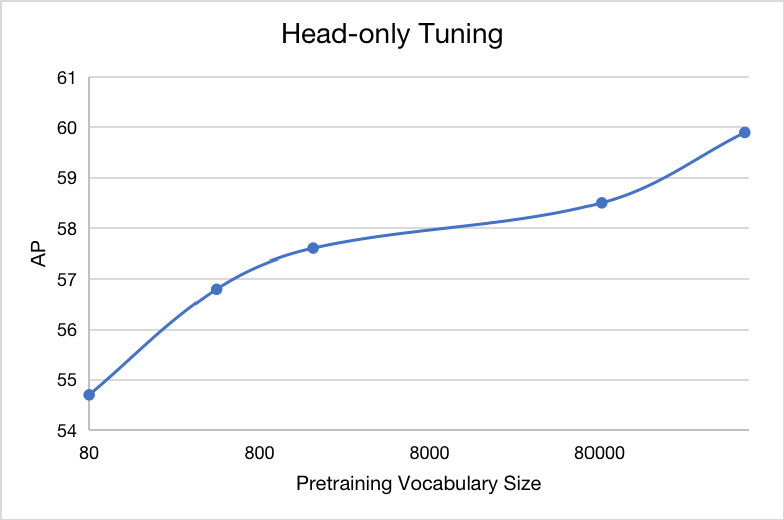}
\includegraphics[width=0.45\textwidth, height=4cm]{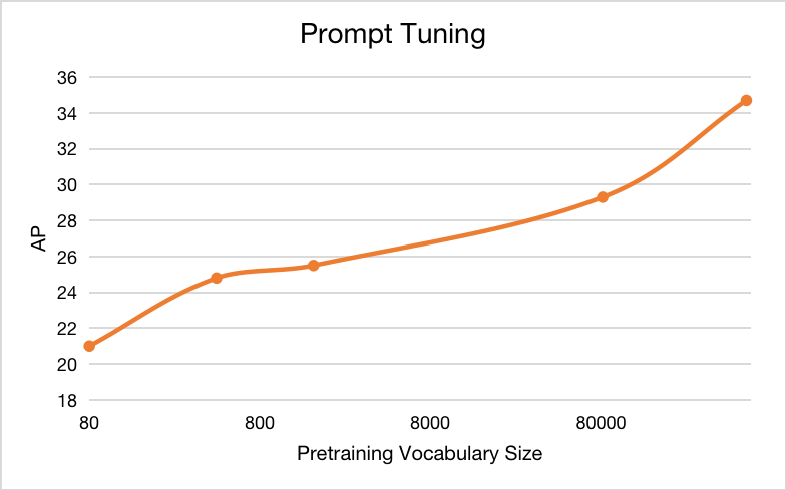}
\caption{Vocabulary size used in pre-training vs. the AP score of fine-tuning on ODinW with head-only and prompt tuning.. X-axis is in log-scale.}
\label{fig:zs}
\end{figure}

Experimental results show that large-scale multi-dataset pre-training is crucial for successful parameter-pretraining (Table~\ref{tab:ELEVATER}). For Head-only FT, the performance drop is reduced from 11.3\% for OmDet-C to only 6.1\% for OmDet. The same trend is observed for Prompt FT, in which the performance drop compared to full-model tuning is reduced from 65.9\% to 45.5\% from OmDet-C to OmDet. Figure~\ref{fig:zs} also visualizes the trend of AP vs. the vocabulary size in pre-training (log-scale). The apparent up-going curve can be observed as more visual concepts are included during pre-training. This suggests that:

(1) multi-dataset pre-training enables the accumulation of a large number of visual concepts, which leads to a stronger backbone that extracts general-purpose visual features (supported by head-only FT results). 

(2) the diversity in language is crucial for successful prompt tuning such that the entire model output can be controlled by the task embedding only (less than 1\% of the parameters of the entire model).

\begin{figure*}[ht!]
\centering
  \includegraphics[width=14cm]{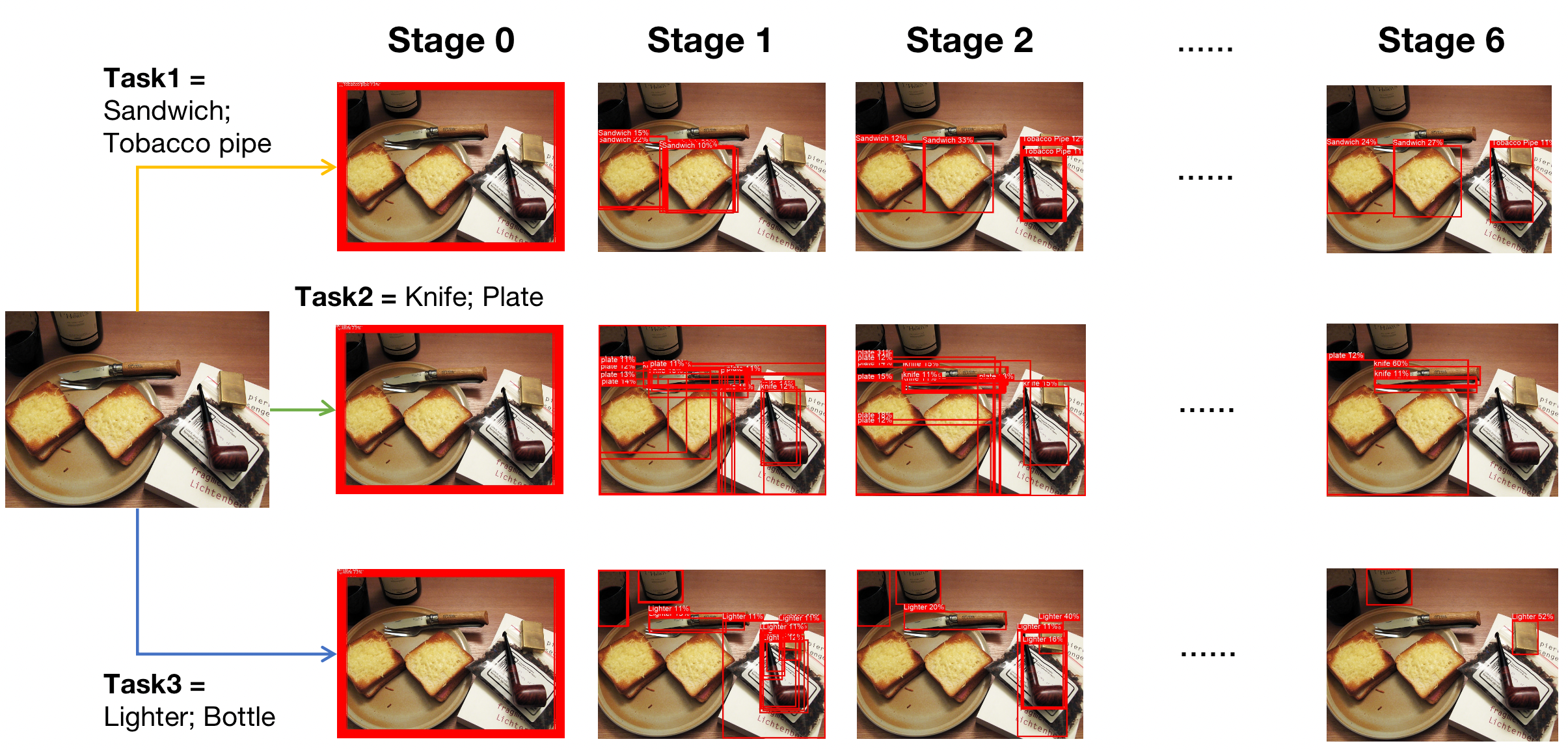}
  \caption{Illustration of language-aware OD, where a single model can generalize (without fine-tuning) to any input tasks on the fly in the form of natural language.}
  \label{fig:viz}
\end{figure*}

\subsection{Visualization of Language-Aware Detection}
Lastly, we conducted qualitative visualizations to showcase the effectiveness of our proposed language-aware object detection model in accurately localizing and labeling objects based on natural language inputs (Figure~\ref{fig:viz}). By inputting different tasks, e.g., [Sandwich, Tobacco Pipe] vs. [Lighter, Bottle], OmDet can dynamically adapt its object localization and classification conditioned on the given task. Figure~\ref{fig:viz} visualizes the intermediate output at each stage of the MDN block. We found that the model learns to place its proposal boxes as the whole image for the initial stage and quickly narrows its focus from the initial whole-image boxes to the objects of interest quickly in 2-3 steps in a top-down search manner. The later stage output continues to refine its output and confidence scores (e.g., with less duplicated bounding boxes and more certain confidence).

\section{Discussions}
\subsection{Inference Speed}
\begin{table}[!h]
\centering
\small
\begin{tabular}{c|c} \hline
         Models &  LVIS FPS (A100) \\\hline
         MDETR        & 0.056         \\
         GLIP-T      &     0.14            \\ 
         OmDet-T*      & 7.74 \\
         OmDet-T       & 8.66 \\
         OmDet-B       & 5.66   \\
         \hline
\end{tabular}
\caption{Inference Speed on LVIS datasets, 12K labels }
\label{tbl:inference_speed}
\end{table}
Table~\ref{tbl:inference_speed} presents a comparison of the inference speeds across different models. Visual grounding MDETR model encodes one class at a time for OD and suffer from slow inferences speed, e.g. 10s/img for MDETR. Also, MDETR uses transformer to fuse image with text, which cannot scale up to multi-scale features due to complexity of self-attention.  In the GLIP method for object detection, objects are identified by combining all their labels into a single descriptive sentence. While this method proves effective in certain contexts, it encounters limitations when applied to datasets with an extensive labels, such as those found in LVIS.  The reason for this slowdown is that creating one big sentence out of many labels creates unnecessary links between the labels, which complicates the detection process and reduces speed. Our MDN deals with fixed-size latent queries, which are independent of visual features. Thus, our method is able to predict many classes with significant speed up with on par or better performance.

\subsection{Different Iterative Fusion}
We have explored the influence of the iterative number on the multi-modal detection network based on iterative fusion in the ODinW datasets. Our investigation involved varying the number of heads in the ConvNext-B architecture, specifically analyzing the performance impact of 1, 3, and 6 heads. The results of these experiments are summarized in Table 11. We observe a substantial improvement when increasing the number of heads from 1 to 3, indicating that additional iterations enhance the network's capability to discern complex patterns in data. This improvement continues, though at a reduced pace, when expanding from 3 to 6 heads. This suggests that while iterative fusion brings benefits in handling complex scenes. 

\begin{table}[ht]
\centering
\small
\begin{tabular}{lccc}
\hline
Models  &  ConvNext-B  & FPS(A100) \\
\hline
1 Head & 47.04  & 7.97 \\
3 Head & 65.60 & 7.49 \\
6 Head & 66.57  & 6.86 \\
\hline
\end{tabular}
\caption{Comparison of AP and Inference Speed with Different Iterative Numbers}
\label{tab:iter_num}
\end{table}

\subsection{Comparison with State-of-the-Art Methods on Open-Vocabulary Benchmarks}
In Table 12, we present a comprehensive quantitative comparison between our proposed model, OmDet, and several state-of-the-art object detection models, namely ViLD~\cite{gu2021open}, CORA~\cite{wu2023cora}, and BARON~\cite{wu2023aligning}. The evaluation is conducted on the widely recognized COCO and LVIS benchmarks, utilizing the open-vocabulary setting as conducted in VILD. The evaluation metrics employed for the comparison include $AP50_{novel}$ and $AP50_{base}$ for COCO, as well as $AP_{r}$ (the AP of rare categories) for LVIS.
The $AP50_{novel}$ score evaluates the model's performance on novel objects, which are not seen during the training phase, while the $AP50_{Base}$ score assesses detection on the base categories, which are present in the training dataset. Additionally, the $AP_{r}$ score provides valuable insights into the model's performance on the 337 rare categories of LVIS that were not part of the training categories. These evaluation metrics collectively offer a comprehensive assessment of the proposed model's effectiveness across different open-vocabulary object detection scenarios and dataset characteristics.
As illustrated in the table, OmDet significantly outperforms the other models across all three metrics. With an $AP50_{novel}$ score of 75.17, an $AP50_{Base}$ score of 70.79, and an $AP_{r}$ of 24.65, OmDet demonstrates superior detection capabilities for both novel and base objects. 

\begin{table}[ht]
\centering
\begin{tabular}{ccccc}
\hline
Model & AP50\_{novel}(COCO) & AP50\_{base}(COCO) & APr(LVIS) \\
\hline
ViLD~ &  27.6 & 59.9 & 16.3 \\
CORA &  43.1 & 60.9 & 28.1 \\
BARON &  42.7 & 54.9 & 23.2 \\
OmDet-B  & 75.2 & 70.8 & 24.7 \\
\hline
\end{tabular}
\caption{Comparative analysis of recent object detection models on OV-COCO and OV-LVIS}
\label{tab:model_performance}
\end{table}

\section{Conclusion}
This work proposes to advance zero/few-shot OD via continual pre-training from a large number of OD datasets by solving the two key technical challenges: \textit{Taxonomy conflict} and \textit{Fore/background inconsistency}. OmDet proposes a novel multimodal detection network that is able to do a fusion of natural language prompts with visual features for language-augmented object detection. Study results confirm the efficacy of OmDet for multi-dataset learning and large-scale pre-training as a foundation model. Our approach OmDet achieved state-of-the-art performance on OV-COCO with a notable $AP50_{novel}$ score of 75.17, $AP50_{base}$ score of 70.79, and an APr score of 24.65 on OV-LVIS, thereby significantly surpassing competing models such as BARON and CORA. We also show that enlarging the vocabulary size via multi-datasets pre-training effectively improves zero/few-shot learning and parameter-efficient fine-tuning. OmDet achieved state-of-the-art performance on 35 downstream tasks from ODinW. Future research will focus on improving OmDet by exploring better text prompt encoding methods and pre-training strategies that will improve zero-shot detection performance and prompt-tuning performance. 
\color{black}

\section*{Acknowledgements}
This research is partially supported by National Key Science and Technology Program of China under grant (2022YFF0902600).

 \bibliographystyle{elsarticle-harv} 
 \bibliography{omdet}

\end{document}